\documentclass[runningheads]{llncs}
\usepackage[T1]{fontenc}
\usepackage{graphicx}
\usepackage{multirow}%
\usepackage{amsmath,amssymb,amsfonts}%
\usepackage{mathrsfs}%
\usepackage[title]{appendix}%
\usepackage[usenames,dvipsnames]{xcolor}
\usepackage{textcomp}%
\usepackage{manyfoot}%
\usepackage{booktabs}%
\usepackage{algorithm}%
\usepackage{algorithmicx}%
\usepackage{algpseudocode}%
\usepackage{listings}%
\usepackage{comment}
\usepackage{siunitx}
\usepackage{setspace}
\usepackage{array}
\usepackage{cite}
\usepackage{hyperref}
\usepackage{color}

\newcommand*\samethanks[1][\value{footnote}]{\footnotemark[#1]}
\begin{document}
\title{Benchmarking Generative AI for Scoring Medical Student Interviews in Objective Structured Clinical Examinations (OSCEs)}
\titlerunning{Benchmarking GenAI for Scoring OSCEs}
\authorrunning{J. Geathers et al.}
%
\author{Jadon Geathers\inst{1}\thanks{These authors contributed equally.}\orcidID{0009-0002-2851-2457} \and
Yann Hicke \inst{1}\samethanks\orcidID{0000-0001-7234-7001} \and
Colleen Chan\inst{2}\orcidID{0000-0001-6323-5673} \and 
Niroop Rajashekar \inst{2}\orcidID{0000-0003-1205-6979} \and 
Sarah Young \inst{1}\orcidID{0009-0002-6203-3986} \and
Justin Sewell \inst{3}\orcidID{0000-0003-4049-2874} \and 
Susannah Cornes \inst{3}\orcidID{0000-0002-3946-2531} \and 
Rene F. Kizilcec \inst{1}\orcidID{0000-0001-6283-5546} \and 
Dennis Shung \inst{2}\orcidID{0000-0001-8226-1842}
}

%
\institute{Cornell University, Ithaca, NY 14850, USA \\ \email{\{jag569,ylh8,sy398,kizilcec\}@cornell.edu} \and
Yale University, New Haven, CT 06520, USA \\
\email{colleenechan@gmail.com, \{niroop.rajashekar,dennis.shung\}@yale.edu}  \and
University of California San Francisco, San Francisco, CA 94143, USA \\
\email{\{justin.sewell,susannah.cornes\}@ucsf.edu}}
\maketitle              
\begin{abstract}
Objective Structured Clinical Examinations (OSCEs) are widely used to assess medical students’ communication skills, but scoring interview-based assessments is time-consuming and potentially subject to human bias. This study explored the potential of large language models (LLMs) to automate OSCE evaluations using the Master Interview Rating Scale (MIRS). We compared the performance of four state-of-the-art LLMs (GPT-4o, Claude 3.5, Llama 3.1, and Gemini 1.5 Pro) in evaluating OSCE transcripts across all 28 items of the MIRS under the conditions of zero-shot, chain-of-thought (CoT), few-shot, and multi-step prompting. The models were benchmarked against a dataset of 10 OSCE cases with 174 expert consensus scores available. Model performance was measured using three accuracy metrics (exact, off-by-one, thresholded). Averaging across all MIRS items and OSCE cases, LLMs performed with low exact accuracy (0.27 to 0.44), and moderate to high off-by-one accuracy (0.67 to 0.87) and thresholded accuracy (0.75 to 0.88). A zero temperature parameter ensured high intra-rater reliability ($\alpha$ = 0.98 for GPT-4o). CoT, few-shot, and multi-step techniques proved valuable when tailored to specific assessment items. The performance was consistent across MIRS items, independent of encounter phases and communication domains. We demonstrated the feasibility of AI-assisted OSCE evaluation and provided benchmarking of multiple LLMs across multiple prompt techniques. Our work provides a baseline performance assessment for LLMs that lays a foundation for future research into automated assessment of clinical communication skills. 

\end{abstract}

%
%
\section{Introduction}\label{sec1}

Effective communication skills are crucial across professional fields, but they are especially critical in healthcare, where they significantly impact patient outcomes and satisfaction \cite{ratna2019importance, bartlett2008impact}. Medical schools face the complex challenge of training and assessing clinical communication skills for thousands of students each year, which are typically assessed through Objective Structured Clinical Examinations (OSCEs) \cite{khan2013objective}. In these assessments, medical students demonstrate their ability to communicate effectively with trained actors serving as standardized patients (SPs), using essential skills like gathering information, building rapport, and showing empathy. The scale of this challenge in medical education is significant: each medical school must evaluate hundreds of student-SP engagements per year, with each OSCE requiring a careful evaluation by humans across multiple communication dimensions.

While video recording technology has made it easier to capture these engagements, the evaluation process remains highly time-intensive and costly \cite{duran2023role}. Moreover, human evaluation of communication skills faces the inherent challenge of inconsistency across different evaluators, which necessitates extensive standardization training \cite{chong2017sights, zimmermann2020standardized}. This creates a bottleneck in providing feedback to students, which is typically delayed by days or weeks after their performance and lacks sufficient detail to help them improve \cite{Uchida2023}.

Large language models (LLMs) offer a promising solution to these assessment challenges by automating both scoring and feedback provision for OSCEs. Beyond just alleviating the time burden on human evaluators, LLMs could enable near-immediate, detailed, and formative feedback to students, which could improve skills development outcomes \cite{harrison2015barriers}. This scoring and feedback automation could fundamentally transform the way medical students train their communication skills: instead of relying on limited formal assessments, medical schools could offer frequent opportunities for deliberate practice with automated feedback under consistent evaluation standards. 

However, automating OSCE assessment presents several challenges. While LLMs can process interview transcripts, they must evaluate complex interpersonal skills that human evaluators typically assess through both verbal and non-verbal cues. Additionally, they must reliably interpret subjective communication assessment rubric items that even human evaluators struggle to score consistently \cite{Fernandez2007, Berg2015, Fluet2022}. Our work tackles these challenges by developing and evaluating an LLM-based system for determining whether LLMs can reliably score medical interviews. As formative feedback hinges on the ability of LLMs to map medical interview transcripts to scores, this represents a critical first step in automating OSCE assessment. 

Our work makes the following contributions:

\begin{enumerate}
    \item We present the first systematic evaluation of LLMs for fully automated OSCE assessment, using a carefully curated dataset of 10 OSCE cases with 174 expert consensus ratings. 
    \item We compare four state-of-the-art LLMs across different assessment tasks, from exact scoring to proficiency band identification, using the Master Interview Rating Scale (MIRS) as our evaluation framework.
    \item We provide a comprehensive analysis of four prompting strategies (zero-shot, chain-of-thought reasoning, few-shot, and multi-step), which we optimized for each assessment criterion.
\end{enumerate}

Our investigation provides insights into the feasibility of automated assessment of interpersonal skills, which is a necessary foundation for developing scalable feedback systems in medical education, and an essential step in contributing to the development of more effective, empathetic, and prepared future healthcare professionals.

\section{Related Work}

\subsubsection{Automated Assessment in Educational Settings.}
Early work on automated essay scoring (AES) demonstrated that hand-engineered text features could approximate human judgments on written prompts \cite{attali2006automated}. Building on these foundations, researchers have integrated machine learning and natural language processing to capture more nuanced attributes of student writing, improving feedback in both secondary and higher education contexts \cite{yang2023my, mcnamara2015hierarchical}. Notably, recent advances in LLMs have led to moderate to high reliability in grading tasks on university exams \cite{floden2024grading, yavuz2025utilizing, wetzler2024grading, dimari2024ai}. While these methods have streamlined grading and feedback in conventional writing or short-answer tasks, they predominantly address static text-based responses rather than interactive, interpersonal communication scenarios that are found in medical interviews.

\subsubsection{Conversation Analysis and Feedback.}
With the rise of conversational AI, studies have explored dialogue-based assessment and support tools. For example, Jain et al. developed multimodal perspective-based dialogue summarization for tutoring sessions \cite{jain2023can}, while Wang et al. introduced systems to provide real-time guidance for teachers \cite{wang2024tutor}, and others have leveraged NLP to measure teaching practices and optimize peer-tutoring interactions \cite{demszky2023m, borchers2024combining}. Although these approaches demonstrate the potential of AI-driven conversation analysis, they primarily focus on pedagogy in K–12 or university settings, where the emphasis is on teaching strategies and student engagement rather than professional communication under clinical constraints.

\subsubsection{AI in Medical Education.}
Recognizing the potential benefits of LLMs for medical training, several studies have examined how models like GPT-4 or specialized medical chatbots perform on clinical reasoning tasks, exam-style questions, and empathy-based prompts \cite{xu2024data, huang2023benchmarking, abd2023large, safranek2023role, luo2024assessing}. Some of these indicate that LLMs can outperform general models in specific contexts \cite{xu2024data}, or even match physician-level empathy in synthetic patient interactions \cite{luo2024assessing}. Yet, despite progress in evaluating clinical knowledge and empathy, few researchers have tackled the complex interpersonal dimensions required for OSCE assessments. Recent studies have explored applying LLMs to portions of OSCE grading workflows. For instance, Shakur et al. \cite{shakur2024large} achieved high concordance ($\kappa$=0.88) between GPT-4 and human graders on a single rubric item (“Did the student summarize the patient’s medical history?”), leveraging transcripts from 2,027 video-recorded OSCEs. Similarly, Jamieson et al. \cite{jamieson2024rubrics} prospectively deployed a GPT-4-based system to grade post-encounter notes, reducing human grading effort by over 90\%. While these works highlight the feasibility and operational benefits of LLM-driven assessment, they each focus on a specific subset of communication competencies—summaries in \cite{shakur2024large} and written post-encounter documentation in \cite{jamieson2024rubrics}—rather than the full spectrum of interpersonal skills evaluated in an OSCE.

In contrast, our study benchmarks multiple prompt strategies (zero-shot, chain-of-thought, few-shot, and multi-step) across all 28 items of the Master Interview Rating Scale \cite{MIRS_UConn}. By examining a comprehensive set of communication behaviors—including empathy, rapport, questioning style, and information-giving—we extend prior work to provide a richer evaluation of how LLMs perform in diverse communication domains. This broader approach not only gauges feasibility but also identifies item-specific challenges and prompt tailoring strategies necessary for robust, automated OSCE assessments in medical education.

\section{Methods}
\begin{figure}[t]
    \centering
    \includegraphics[width=\linewidth]{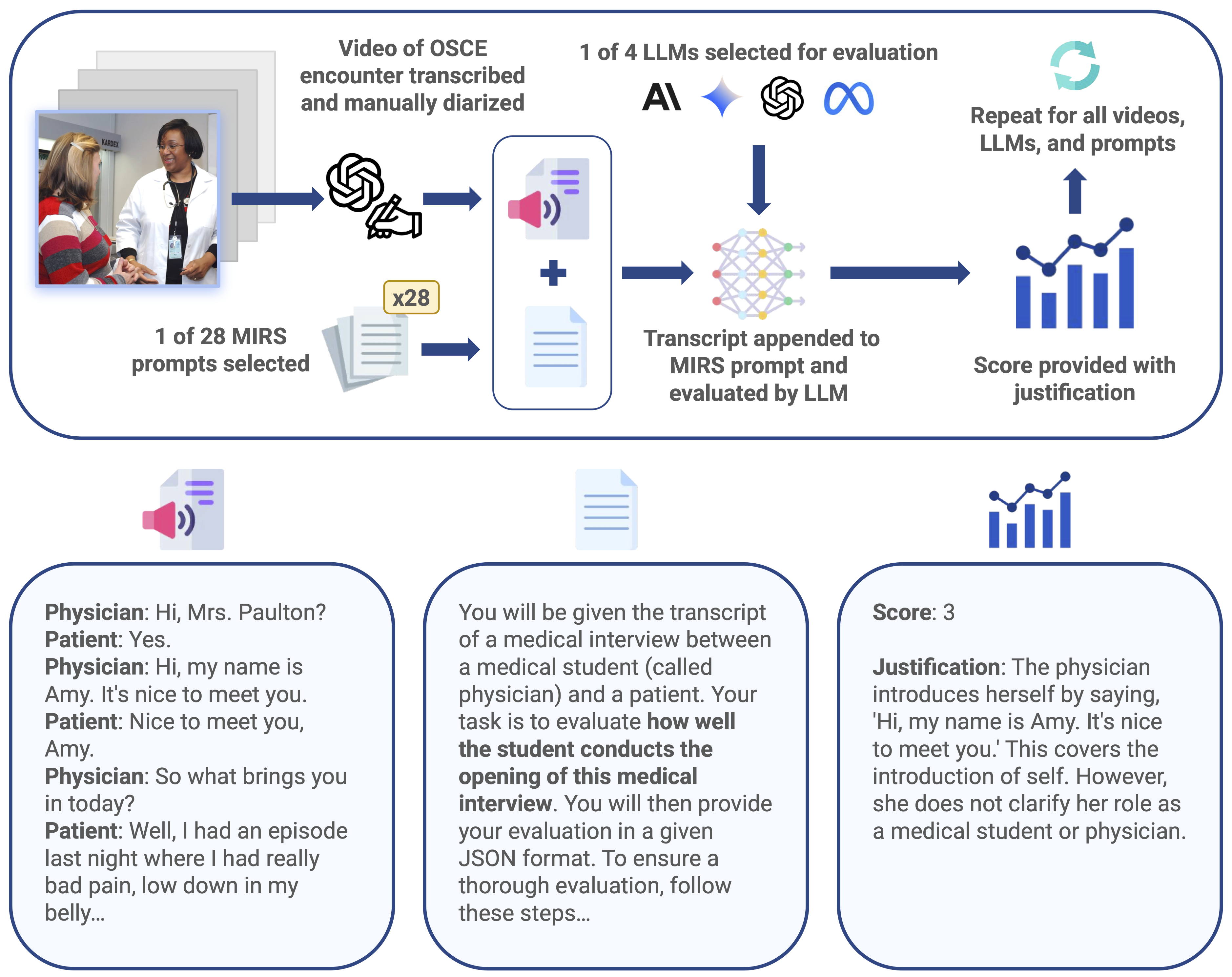}
    \caption{The overall flow of the evaluation process (top), along with examples of an annotated transcript, MIRS item prompt, and score/justification pair (bottom). OSCE video transcripts are appended to MIRS item prompts, which are passed into LLMs for scoring and justification.}
    \label{fig:benchmark-flow}
\end{figure}
\subsection{Dataset}
This benchmarking study analyzed a dataset containing 10 unique OSCE video recordings, each featuring a different clinical scenario between a health professions student and a standardized patient. The cases capture authentic clinical interactions through videos ranging from 7 to 30 minutes in duration. This video dataset consists of three distinct categories of clinical engagements: four medical history-taking cases (two focusing on groin pain; two examining left chest pain), three behavioral counseling cases (smoking cessation, exercise, nutrition), and three dental cases (tooth pain evaluation, gum pain assessment, smoking cessation counseling). Expert evaluators from the University of Connecticut provided consensus scores on the MIRS rubric for these encounters, yielding 174 individual scored rubric items in the 10 cases (some MIRS items were not applicable to certain types of cases, resulting in fewer than the theoretical maximum of 280 scored items) \cite{MIRS_UConn}. From each video, we extracted the audio and transcribed the resulting MP3 file using Whisper \cite{Whisper_OpenAI}. The dialogue between the student physician and standardized patient was then diarized through manual annotation by our team (Figure \ref{fig:benchmark-flow}). 

\subsection{Evaluation: Master Interview Rating Scale}
We based our evaluation on the Master Interview Rating Scale (MIRS), a validated instrument for assessing medical communication skills \cite{osullivan2008development, baldwin2017delivery}. The MIRS comprises 28 items, each rated on a 5-point scale with three labeled anchor statements, assessing various aspects of the medical interview including questioning skills, interview organization, and patient inclusion. Expert consensus scores on the MIRS rubric provided by the University of Connecticut yielded 174 scoring data points across all 10 cases. 

Guidance on how to score each MIRS item was integrated from the University of Connecticut MIRS rubric \cite{MIRS_UConn} and supplemented with examples and contextual notes from the University of Tennessee MIRS rubric \cite{uthsc_mirs}. Although the MIRS is scored on a 5-point scale, it only has labeled anchor statements for scores of 1 (lowest score), 3 (mid-point), and 5 (highest score). To give clear scoring instructions to the LLM, our team wrote anchor statements for scores of 2 and 4, with medical education subject matter experts on our team validating the language and suitability of the scoring criteria. Most of the MIRS items (26 out of 28) can be scored with LLMs based on the text alone by using automatically generated transcripts. The remaining two MIRS items–"Pacing of Interview" and "Non-Verbal Facilitation Skills"–required a multimodal evaluation strategy because they could not be assessed through text transcripts alone. Most of the analysis in our work centers of the 26 verbal items, but we also report findings for the two non-verbal items using a multimodal analysis.

\begin{figure}
    \centering
    \includegraphics[width=\linewidth]{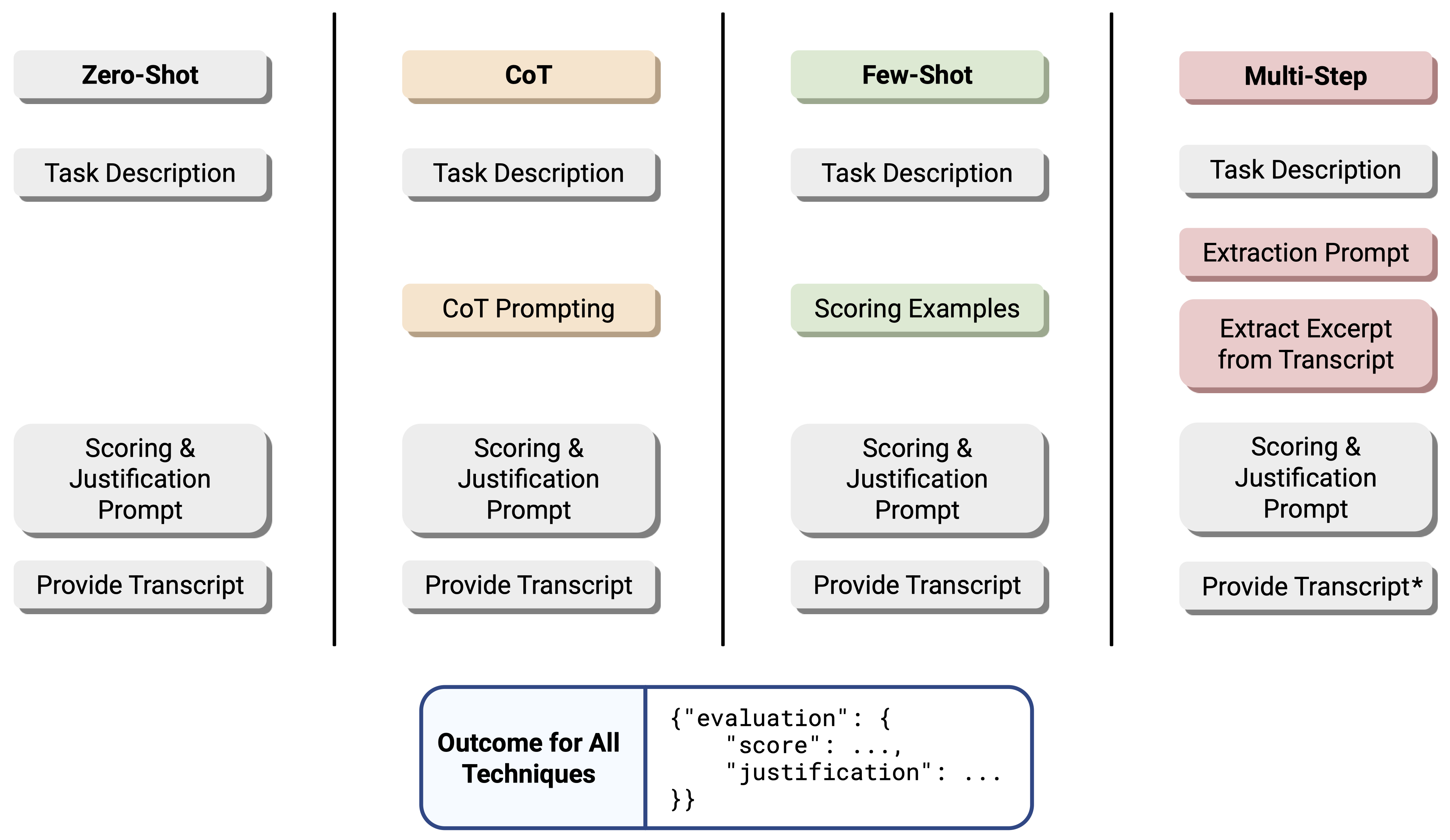}
    \caption{Structure of the steps involved for each prompt, depending on the prompting technique, with one such prompt for each MIRS rubric item. In multi-step prompting, the “Provide Transcript” step uses the extracted excerpt.}
    \label{fig:prompting}
\end{figure}

\subsection{LLMs and Prompting Techniques}
We tested the performance of four state-of-the-art language models in this study: GPT-4o (OpenAI) \cite{GPT4o_2024}, Claude 3.5 (Anthropic) \cite{Claude3.5_2024}, Llama 3.1 (Meta) \cite{LLaMA_v3p1_405B_Instruct}, and Gemini 1.5 Pro (Google) \cite{Gemini1.5_Pro_Exp_0801}. All models were configured with a temperature of 0 for the most deterministic model responses. Additionally, we explored four prompting techniques to optimize model performance:

\begin{itemize}
    \item \textbf{Zero-shot.} We used this technique to establish our baseline performance, using only our adapted MIRS rubric for scoring.
    \item \textbf{Chain-of-Thought (CoT) Reasoning.} The model was prompted to generate a structured list of key statements from the transcript that were relevant to the assessed item before scoring \cite{wei2022chain}. This intermediate step encouraged the model to articulate its reasoning, with the list used to justify score selections.
    \item \textbf{Few-shot.} Up to 5 examples of relevant statements from physician-patient interactions were provided to guide the models’ reasoning \cite{wang2020generalizing}. These were paired with an appropriate score or justification, as provided by the University of Tennessee’s rubric.
    \item \textbf{Multi-step.}  In this approach, the model was first prompted to extract relevant excerpts from the transcript before using these excerpts to inform scoring. This method was inspired by prior work on multi-step prompting \cite{10.1145/3583780.3615265}.
\end{itemize}

To prompt the model to conduct OSCE assessments, we constructed individual prompts for each item of the MIRS rubric. While prompts differed by prompting technique, they maintained a consistent underlying structure within each approach. As shown in Figure~\ref{fig:prompting}, each prompt began with a task description that outlined the evaluation scenario and objectives. The subsequent structure then diverged according to the prompting technique employed. For CoT reasoning, we requested the model to establish a list of key statements identified in the dialogue to help motivate evaluation. In the case of few-shot prompting, we included examples of relevant excerpts for the rubric item and their assigned scores. All approaches then provided the scoring rubric, requesting the model to evaluate the medical student's performance based on the scoring criteria. The model was prompted to provide a direct, evidence-based justification for its score selection. Lastly, we provided the model with the transcript, or the relevant transcript excerpt if using multi-step prompting. All prompts are available on GitHub\footnote{https://github.com/YannHicke/MedEd/tree/main/code/evaluation/prompts}.

\subsection{Evaluation Metrics}
We evaluated model performance relative to expert consensus answers provided by medical educators at the University of Connecticut\cite{MIRS_UConn}, who authored the OSCE cases and assessed the students' performances according to the MIRS rubric. We used three accuracy metrics for evaluation, each representing a different level of scoring leniency:

\begin{itemize}
    \item \textbf{Exact Accuracy (Conservative).} We measured the exact agreement between the expert consensus and model scores, providing the strictest evaluation criterion. Exact accuracy is relevant for assessing the models' ability to exactly match human judgment for precise assessment of communication competencies in OSCEs.
    
    \item \textbf{Off-by-one Accuracy (Moderate).} We measured approximate agreement within one point (above or below) of the expert consensus scores. This recognizes the subjective variability inherent to clinical evaluations among human raters \cite{mclaughlin2009effect, hope2015examiners}. The metric provided a less strict measure of accuracy, but it reflects a practical balance between strictness and flexibility in evaluating performance. 

    \item \textbf{Thresholded Accuracy (Lenient).} We measured discrepancies between the consensus and the model scores by creating two score buckets: one bucket contained scores 1 and 2 (below proficiency threshold) and the other bucket contained scores 3, 4, and 5 (meets or exceeds proficiency, as suggested by medical experts on our team). This metric assessed the models' ability to distinguish between broader proficiency levels, aiding practitioners in identifying students who may require additional support.
\end{itemize}

\subsection{Experimental Details}
The complete evaluation flow used to produce scores and justifications is illustrated in Figure~\ref{fig:benchmark-flow}. All models were configured with a temperature of 0, ensuring that they always selected the most probable next token when generating responses. This setting made the responses largely deterministic, which removed the need for repeated trials due to consistent, nonrandom token selection. As a temperature of 0 can occasionally produce slight variations in output (e.g., due to tie-breaking between equally probable tokens), we conducted an intra-rater reliability test for GPT-4o using the baseline zero-shot prompting strategy to confirm the consistency of the model’s scoring.

For this reliability test, we obtained five independent evaluations for each case and treated each evaluation as an independent rater. Using Krippendorff's alpha, a reliability coefficient suited for ordinal data, we measured the agreement between the model's scores across trials. Krippendorff's alpha accounts for the degree of disagreement, penalizing larger discrepancies more heavily (e.g., a difference between 1 and 4 is penalized more than between 2 and 3). The coefficient ranges from -1 (indicating systematic disagreement) to 1 (indicating perfect agreement). The test yielded a Krippendorff's alpha of 0.98 for GPT-4o, indicating excellent internal consistency.

Beyond the reliability test, we demonstrated the overall performance of each model (Claude, Gemini, GPT, Llama) across prompting techniques on the three accuracy metrics. This involved conducting an analysis where we selected the optimal prompting technique for each MIRS item to see whether adapting the prompting strategy to item characteristics led to improvements in model agreement with the consensus scores. We also assessed the performance of all models on each relevant MIRS item by evaluating off-by-one accuracy for the baseline approach. In this analysis, the items were categorized by their occurrence during the encounter and their communication domain. 

Finally, we discuss early benchmarking findings of multimodal models, which incorporate audiovisual data, for evaluating MIRS items that are not textually represented in a direct transcript. This involved providing the dataset videos to Gemini 1.5 Pro with our prompts and evaluating the video and audio directly rather than using a transcript.

\section{Results}
\subsection{Overall Model Performance}
\begin{figure}
    \centering
    \includegraphics[width=\linewidth]{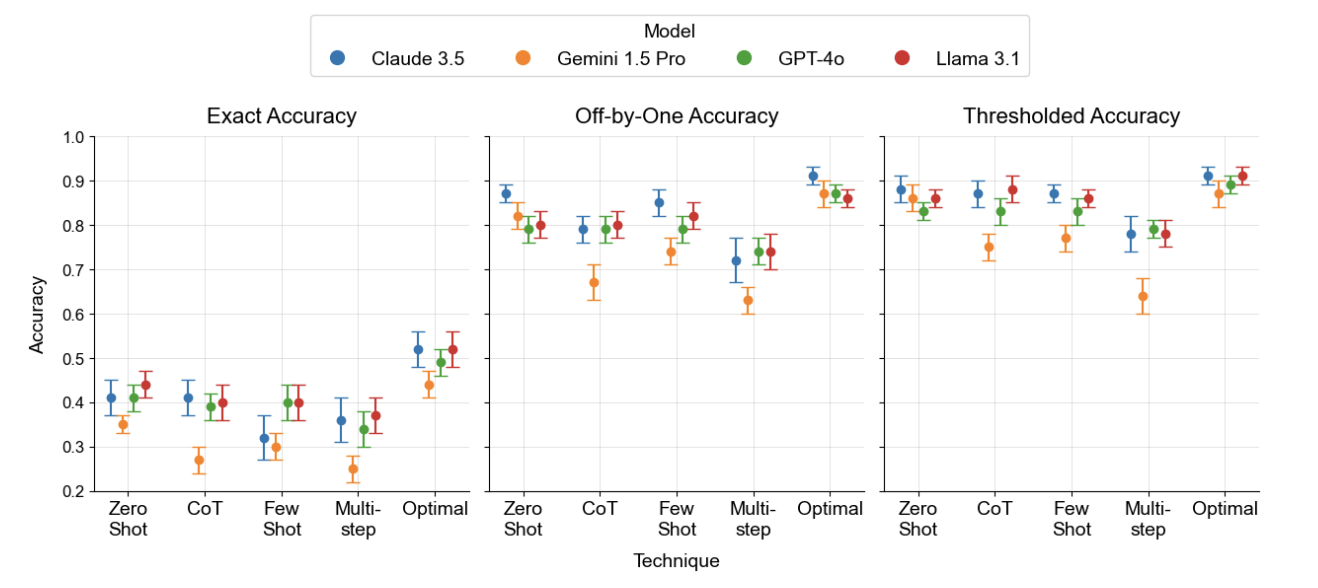}
    \caption{Average performance of each model using different prompting techniques and measured with different accuracy metrics. Error bars represent standard errors calculated across the 10 OSCE cases, where each case's accuracy is first computed as the mean of all applicable MIRS items for that case.
}
    \label{fig:results}
\end{figure}

Figure~\ref{fig:results} illustrates LLM performance across prompting techniques using three measures of accuracy, aggregating over the MIRS items. While exact accuracy was low (0.27 to 0.52), off-by-one accuracy (0.67 to 0.91) and thresholded accuracy (0.75 to 0.91) were moderate to high. Surprisingly, CoT, few-shot, and multi-step techniques did not improve performance over the zero-shot baseline. However, selecting the optimal prompting technique for each MIRS item improved performance, indicating that adjusting the prompting technique to the rubric item may be beneficial. We note this represents a theoretical upper bound since optimal techniques were selected on the same dataset used for evaluation. While this approach helps identify which prompting strategies show promise for different rubric items, future work should validate these findings using proper cross-validation.

Although the exact accuracy was relatively low (0.52 for Claude), the models’ ability to differentiate performance levels (thresholded accuracy 0.83-0.88 for zero-shot) aligns with practical applications in OSCE evaluations, as broad proficiency levels are often more informative than exact scores. This suggests LLMs are better used as complementary evaluation and proficiency detection tools rather than standalone replacements. All augmentative prompting techniques (CoT, few-shot, and multi-step) either minimally impacted or decreased accuracy compared to the zero-shot baseline. Declines in accuracy were particularly notable with few-shot prompting for Gemini and universal with multi-step prompting. 

\subsection{Performance by MIRS Items}
\begin{figure}
    \centering
    \includegraphics[width=0.9 \linewidth]{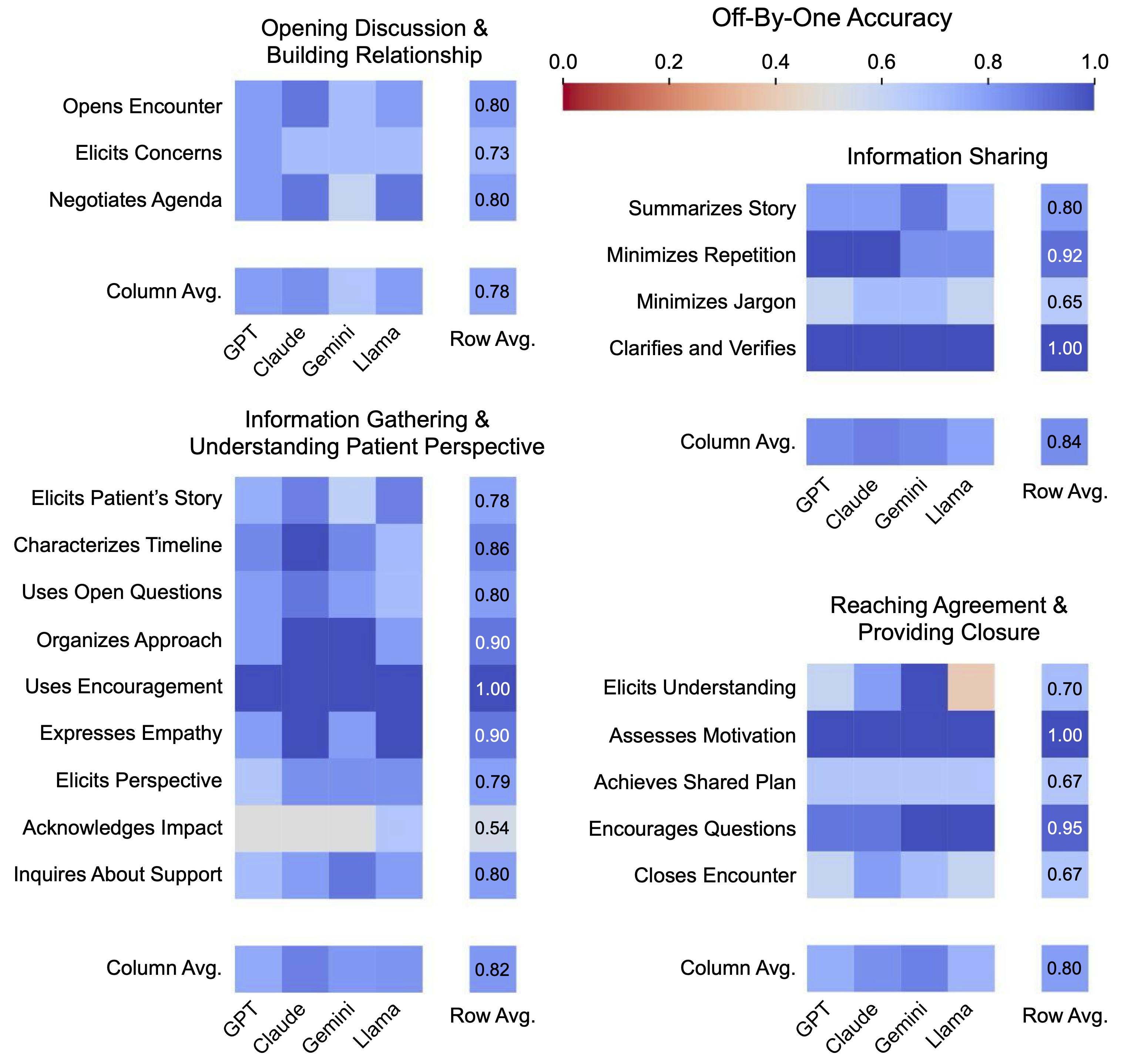}
    \caption{Heatmap of the off-by-one accuracy for each model and MIRS item. While we analyzed a total of 26 text-based MIRS items, this figure displays results for the 21 items where data points were available across all cases. Opening the discussion and building the relationship correspond to the beginning of the visit. Information gathering and understanding of the patient's perspective occur primarily during the middle of the visit. Information sharing occurs throughout the visit. Reaching an agreement and providing closure correspond to the end of the visit.}
    \label{fig:flow}
\end{figure}
We analyzed off-by-one accuracy for each MIRS item across the four models (Claude, Gemini, GPT, Llama) using the zero-shot baseline approach, grouping items into four temporal, skill-based phases\cite{joyce2010use}. 
We selected off-by-one accuracy because it provided a balanced evaluation approach between lenient thresholded accuracy and conservative exact accuracy, though alternative accuracy metrics showed similar patterns for underperforming MIRS items. Most items were scored with an accuracy of 0.8 or higher, indicating the models’ alignment with consensus scores across a variety of communication assessment tasks.

While models demonstrated similar patterns of accuracy across encounter phases, specific items consistently challenged all models (e.g., “acknowledges impact,” “minimizes jargon,” “achieves shared plan”) while others performed uniformly well (e.g., “clarifies and verifies,” “assesses motivation,” “uses encouragement”). The variable performance across individual items may be attributed to several factors. First, performance could be influenced by the clarity of item definitions, the frequency of items in transcripts, and the quality of provided examples. Second, some items were not present in all transcripts, resulting in smaller evaluation datasets compared to items that occurred consistently across cases. As encounter phases alone evoke no clear patterns in accuracy, these findings underscore a need to fine-tune the models on patient-physician encounter data.

\subsection{Multimodal Performance}
We used a multimodal model, Gemini 1.5 Pro, to score the two MIRS items requiring the assessment of non-verbal elements in patient-physician encounters. Performance was notably poor, systematically disagreeing with human raters (Krippendorff’s alpha of -0.47). This failure to effectively reason on the basis of the provided audiovisual data may have arisen because the case videos often capture the expressions and gestures of one speaker at a time rather than capturing the engagement between both speakers simultaneously. The systematic disagreement with human raters suggests fundamental limitations in the ability of current multimodal models to interpret the nuanced non-verbal communication elements central to clinical encounters.

\section{Discussion and Future Directions}
The MIRS rubric assesses diverse communication skills. Some items assess localized elements (such as the encounter opening) and focus on specific phrases, while others assess recurring elements (like empathy statements) or holistic interaction qualities. While clinical skills assessment rubrics like the MIRS are designed with anchor statements to improve scoring reliability, this variability in items–along with rater characteristics, rubric complexity, and the length of assessment items–can undermine inter-rater reliability\cite{bernardin2016rater, vu1992standardized}. These interpretation challenges are similarly reflected in LLM-based assessment, where unclear rubric language can lead models to misapply the scoring criteria.

To address these challenges, we explored various prompting techniques, and found that uniformly applied techniques led to performance declines, revealing limitations in common approaches. Few-shot prompting, despite its wide use, led models to overemphasize model language rather than evaluate the overall communication quality. Similarly, multi-step prompting proved ineffective, as the models failed to identify all relevant transcript elements, especially for MIRS items requiring consideration of the full transcript. In contrast, dynamically selecting the optimal technique for each MIRS item showed promise, highlighting the potential for performance enhancement through tailored approaches. The poor performance of multimodal models on non-verbal assessment items highlights a significant gap in current AI capabilities. Future work should explore specialized architectures for clinical encounter assessments that can better integrate sequences of non-verbal cues with verbal content.

While our work establishes baseline performance using standard LLMs with different prompting techniques, several limitations remain. LLMs performed better on rubric items with concrete behaviors than those requiring nuanced judgment, suggesting behavioral anchors may be more suitable for automated assessment. During development, we observed that subtle prompt variations influenced scoring outcomes despite zero-temperature settings, and LLMs may introduce biases from dominant communication patterns in their training data. Fine-tuning models on domain-specific OSCE evaluation data could enhance performance, particularly for challenging MIRS items with lower agreement with expert consensus. Additionally, since our benchmark uses data from a single institution and OSCE implementation varies across medical schools, cross-institutional validation is an important next step for generalizability.

Lastly, while this study focused on MIRS scoring agreement between LLMs and human raters, LLMs can also provide written feedback, which is an important component of OSCE evaluations. Future work should consider how to align feedback with the learning objectives of clinical skills curricula and students’ personal development goals while providing justifiable, targeted areas for improvement. This entails improving prompt design, contextualization, and review by medical students and educators. We encourage researchers to build on our benchmark by developing prompts that guide models toward realistic assessments without unnecessary leniency or bias.

\section{Conclusion}
In this benchmarking study, we demonstrated the use of LLMs in automating the evaluation of Objective Structured Clinical Examinations (OSCEs). Models exhibited high agreement with human evaluators in identifying students needing support (low-performing scores of 1-2) across assessment items using zero-shot prompting. CoT, few-shot, and multi-step proved useful when tailored to specific assessment items. Overall, LLMs show promise for effective automation and application to OSCE evaluation, but challenges remain in refining the prompt design and ensuring consistent numerical scores and useful feedback. These findings contribute to the broader understanding of how LLMs can support the assessment of complex human interactions in educational settings. We provide our prompts and evaluation approach as an initial benchmark and invite further improvements from the research community in developing more robust and reliable automated OSCE assessment systems that can enhance medical education.

\bibliography{bibliography.bib}
\end{document}